\documentclass[10pt,twocolumn,letterpaper]{article}

\usepackage{iccv}
\usepackage{times}
\usepackage{epsfig}
\usepackage{graphicx}
\usepackage{amsmath}
\usepackage{amssymb}
\usepackage{wrapfig}
\usepackage{subfiles}
\usepackage[vlined, ruled]{algorithm2e}
\usepackage{multirow}
\usepackage{subcaption}
\usepackage{float}
\usepackage{authblk}

\usepackage{hyperref}
\hypersetup{breaklinks=true,colorlinks}

\iccvfinalcopy 

\begin{document}

\title{BID-NeRF: RGB-D image pose estimation with inverted Neural Radiance Fields}

\author[1,2]{Ágoston István Csehi}
\author[1]{Csaba Máté Józsa}

\affil[1]{Nokia Bell Labs, 36-42 Bókay János str., Budapest 1083, Hungary}
\affil[2]{Budapest University of Technology and Economics, 3
Műegyetem rkp., Budapest 1111, Hungary \authorcr
{\tt\small Email: csehi@edu.bme.hu, csaba.jozsa@nokia-bell-labs.com}}


\maketitle
\ificcvfinal\thispagestyle{empty}\fi

\begin{abstract}
We  aim to improve the Inverted Neural Radiance Fields (iNeRF) algorithm which defines the image pose estimation problem as a NeRF based iterative linear optimization.
NeRFs are novel neural space representation models that can synthesize photorealistic novel views of real-world scenes or objects.
Our contributions are as follows:
we extend the localization optimization objective with a depth-based loss function,
we introduce a multi-image based loss function where a sequence of images with known relative poses are used without increasing the computational complexity, 
we omit hierarchical sampling during volumetric rendering, meaning only the coarse model is used for pose estimation, and we show that by extending the sampling interval convergence can be achieved even for higher initial pose estimate errors. 
With the proposed modifications the convergence speed is significantly improved, and the basin of convergence is substantially extended.

\end{abstract}


\section{Introduction}
Image pose estimation has been the focus of numerous research articles.
A robust and efficient solution to this problem is particularly useful for extended reality, robotics, and Simultaneous Localization and Mapping (SLAM) applications.  
There are several competing paradigms, like the feature-based approach composed of: interest point detection, matching, filtering, and pose estimation steps \cite{Campos_2021, detone2018superpoint, sarlin2020superglue}; or the render-and-compare method which tries to reduce the visual differences between a rendered and an observed image \cite{inerf, tremblay2018deep, lin2023parallel}. The render-and-compare paradigm enables the solution of pose estimation up to pixel-level accuracy based on an accurate model of the environment. In our solution, we follow the latter, which is convenient if the reconstruction of the environment's three-dimensional structure is feasible. 

In recent years, a novel space representation approach gained disruptive popularity using NeRFs \cite{nerf}. 
In essence, NeRFs can be seen as implicit functions that define the scene’s surface, depth and appearance properties, making them particularly suited for complex and detailed scene reconstruction.
The advantage of NeRFs is that training can be done based on a sparse set of two-dimensional RGB images with known relative poses. 
The relative pose information required for the training can be computed with the help of structure-from-motion or visual(-inertial) odometry algorithms \cite{Schonberger_2016_CVPR, Qin_2018, engel2018direct, Campos_2021}. 

Improving the training and rendering speed \cite{hu2022efficientnerf,mueller2022instant, yu2021plenoxels, kerbl3Dgaussians, kurz2022adanerf}, the modeling accuracy \cite{barron2023zipnerf, meuleman2023localrf}, and investigating additional applications of NeRFs shows high interest in the research community. 
NeRFs have potential applicability in depth maps generation \cite{dsnerf}, can be used in navigation \cite{adamkiewicz2022vision}, \cite{xie2023navinerf}, localization \cite{maggio2022locnerf, inerf} and six degrees of freedom orientation estimation \cite{li2022nerf}. 

The problem we aim to solve is to better approximate a captured image's absolute pose with respect to a reference NeRF model representing the scene.

\section{Literature Review}
NeRFs \cite{nerf} are neural networks that represent 3D scenes as continuous functions, mapping 3D coordinates to their corresponding scene appearance properties, such as RGB color and volumetric density value. This continuous representation distinguishes them from most traditional 3D models, which often rely on discrete voxels or point clouds.
For rendering RGB images, the final pixel color values are calculated based on differentiable volume rendering \cite{niemeyer2020differentiable}. 
Rays are cast from the virtual camera's center through each pixel. 
Points are sampled along the rays where the model is evaluated. 
With the given volumetric density predictions, the accumulated transmittance can be formulated for each point, which serves as the probability that the ray reaches that point without hitting any other obstacle. With the transmittance as weights, the RGB predictions can be averaged, yielding the final RGB values for each pixel.

In order to improve sampling efficiency by avoiding the oversampling of empty spaces,  in \cite{nerf} the concept of hierarchical sampling was introduced, where two neural networks are used simultaneously. First, a coarse model is used, where sample points are chosen with a uniform distribution along each ray. The inferred volumetric density values are then used to better estimate where the object's surface will intersect the ray. Based on this information a second set of locations is sampled using inverse transform sampling for the fine model, choosing points closer to the object's surface. In the original NeRF implementation, the structural complexity of these neural networks is the same, but there are twice as many sample points along the rays during the evaluation of the fine model.

In \cite{inerf} it was shown, that a trained NeRF model can be used for pose estimation as the result of an optimization problem. 
NeRFs can render virtual images for any given virtual camera pose. 
A pose is defined as a member of the Special Euclidean group SE(3), which has 6 degrees of freedom. 
iNeRF inverts the mapping between pose and image, utilizing the differentiability property of the volumetric rendering pipeline. 
To estimate the pose of an observed image iNeRF renders pixels from an initial pose and compares them with the corresponding pixels of the observed image resulting in the intensity loss. 
The gradients of the intensity loss function with respect to the initial pose are driving first-order optimization algorithms in an iterative way. 
The algorithm was tested on both real-life and synthetic datasets and achieved convergence only if the initial pose was already close to the pose of the observed image. Real applications of this approach are only plausible if both the range and speed of convergence can be further improved.

The beneficial effect of depth supervision for the training of NeRFs was first shown in DS-NeRF \cite{dsnerf}. 
Faster training and a more accurate space/structure representation was achieved by extending the original loss function with a sparse depth-based loss term.
Many structural artifacts that are present in the case of NeRF models trained without depth supervision can be removed, making possible the use of structural information for pose estimation.
Reference depth information can be gathered by using structure-from-motion, depth prediction algorithms or by sensorial means. 
On the other hand, depth values can be easily predicted for the rendered images, without any further computational requirements. 
Our method extends the original pose estimation objective with depth-based loss terms, therefore we utilize depth information during localization as well.
To this end, it is beneficial to train NeRF models that have accurate structure.
In our experiments, all the NeRF models were trained with depth supervision.


Recently, the iNeRF algorithm was enhanced in \cite{lin2023parallel, maggio2022locnerf} by formulating the camera pose estimation in a particle filter framework where the weights/goodness of the particles are defined as the loss between a rendered and an observed image as was introduced originally in \cite{inerf}.
Multiple camera pose hypotheses are simultaneously optimized using a Monte Carlo sampling method.
It was also shown that decoupling the translational and rotational components of the camera pose parameterization is beneficial for the estimation process.
The suggested modifications resulted in significant improvements in optimization robustness and speed compared to \cite{inerf}.
Our contributions focus on creating a better localization objective, therefore they can be combined with the contributions of  \cite{lin2023parallel} and \cite{maggio2022locnerf}.



\section{Methodology}
\label{methodology}
To use NeRFs for image pose estimation in real-time applications, the performance and robustness of the algorithm have to be improved while storage requirements and inference time have to be reduced. To achieve this, we introduce three major modifications compared to previous research \cite{inerf, lin2023parallel}:
we extend the localization optimization objective with a depth-based loss function,
we sample reference pixels from multiple view directions based on a moving window of previous frames and their relative transformation estimations,
we omit hierarchical sampling during volumetric rendering, meaning only the coarse model is used for pose estimation. Our algorithm is formulated in \ref{alg:one}, which we call BID-NeRF, as in Bundle-adjusted Inverted Depth supervised Neural Radiance Fields.

The iNeRF's original problem formulation can be seen in Eq. \ref{eq:inerf_problem} where \(\hat{c}_{r}\) is the rendered color for the ray \(r\), and \(c_{r}\) is the corresponding pixel from the reference image. 
Each ray \(r\) is transformed to the reference coordinate frame from the camera's local coordinate frame with the actual prediction of the camera's \(T\) pose.
\begin{align}
    T^{*}&=\arg \min_{T \in SE(3)} \overbrace{\frac{1}{|R|}\sum_{r\in R\subset I_{rgb}}\left \| \hat{c}_{r}-c_{r} \right \|_{2}^{2}}^{L_{rgb}(T|I_{rgb},\Theta)}
    \label{eq:inerf_problem} \\
    \delta^*&=\arg \min_{\delta\in\mathfrak{se}(3)} L_{rgb}(T_{init}\oplus \delta|I,\Theta)
    \label{eq:final_opt}
\end{align}

While several optimization algorithms exist, we decided to use the Adam optimizer, introduced in \cite{kingma2017adam}. 
The updates to the image poses in \(\mathfrak{se}(3)\) are calculated in the local tangent space of SE(3) \cite{lie_theory}. 
Eq. \ref{eq:final_opt} shows the ideal solution to the formulated problem, where \(T^{*}\) is the globally optimal transformation and \(\delta\) is the inverse of the initial error represented in the local tangent space. 
If this inverse vector was known, we could simply add it to the initial pose estimate to get the optimal solution.
Instead of directly searching for the optimal vector though, we iteratively update our pose estimate as shown in Alg. \ref{alg:one}. The \(\oplus\) operator adds a vector from a Lie algebra to an element from its corresponding Lie group by projecting the vector to a new group element and adding them together through the group's composition operation \cite{lie_theory}.

\subsection{Depth information}
DS-NeRF \cite{dsnerf} showed how depth supervision improves the convergence rate for training NeRFs, by helping the model reconstruct the structure of the environment. 
This way the learned space representation will contain fewer structural noise or artifacts caused by the errors of volumetric density predictions. 
The appearance of such artifacts can be best described as the result of the shape-radiance ambiguity in NeRFs, explained in \cite{zhang2020nerf}. This effect can be reduced by explicitly introducing more constraints in the learning process.

Depth supervision can be used as such a constraint, improving the precision of the predicted depth maps. 
Depth information can also be utilized during pose estimation if depth information is available. 
Depth maps are free of textures and lighting differences. Therefore, they have smoother image-space gradients compared to RGB images. 
\textit{Our hypothesis is that during pose estimation both convergence speed and basin of convergence benefit from the incorporation of depth information.}
\begin{align}
    \hat{d}(r)&=\sum_{i=1}^{N}T_i(1-e^{-\sigma_i(t_{i+1}-t_i)})t_i
    \label{eq:depth_prediction} \\
    &\quad\text{where } T_i=e^{-\sum_{j=1}^{i-1}\sigma_j(t_{j+1}-t_j)} \nonumber \\
    L_{d}(T|I_{d},\Theta)&=\frac{1}{|D|}\sum_{d\in D\subset I_{d}}\left \| \hat{d}-d \right \|_{2}^{2} \\
    L(T|I,\Theta)&=L_{rgb}(T|I_{rgb},\Theta)+\lambda_{d}\cdot L_{d}(T|I_{d},\Theta)
    \label{eq:depth_and_rgb_loss}
\end{align}

Depth predictions $\hat{d}$ for rays $r$ are calculated as shown in Eq. \ref{eq:depth_prediction}, where $t_i$ denotes the distance of the $i$-th sample point from the camera and $\sigma_i$ is the model's density prediction for that point.
Eq. \ref{eq:depth_and_rgb_loss} shows how depth predictions can be added to the original RGB loss, where $\lambda_{d}$ is a new, application-dependent hyperparameter.
In our tests, we used $\lambda_{d}=1$, which our experiments proved to be a reasonable choice.
The added loss term $L_{depth}$ is the mean-squared error between the predictions and the ground-truth depth values $d$, which is similar to the intensity loss term $L_{rgb}$ as shown in Eq. \ref{eq:inerf_problem}. 
The depth predictions do not increase the computational complexity of the volumetric rendering.

If dense depth predictions for every RGB pixel cannot be obtained, unlike with RGB-D cameras, other methods can be used as well. The depth supervision during pose estimation also works with sparse depth information, as shown in \cite{dsnerf}. These can be obtained by off-the-shelf structure from motion algorithms such as COLMAP \cite{Schonberger_2016_CVPR}. Another approach would be to use lidar sensors and render separate virtual rays using volumetric rendering to match the sensor's field of view.
\subsection{Multiple images}

\begin{figure}
\centering
\includegraphics[width=0.8\linewidth]{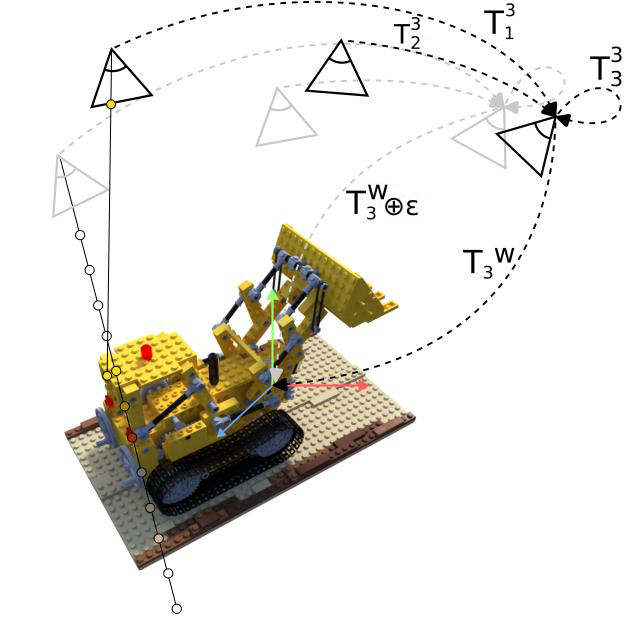}
\caption{Simultaneous localization of multiple images, estimating only the last camera pose.}
\label{fig:multi}
\end{figure}

Having initial pose estimates for each input image obtained by a visual(-inertial) odometry algorithm, we can assume that the errors of their relative transformations are significantly lower compared to the errors of their absolute transformations. 
Errors in state estimation solutions usually come from an accumulated drift, which is ideally infinitesimal for only a few steps. With this assumption, a moving window of sequential images can be localized simultaneously. The concept of image sequence localization with known relative pose transformation is shown in Fig. \ref{fig:multi}.

\begin{equation}
    \delta^{*}=\arg\min_{\delta\in\mathfrak{se}(3)}\frac{1}{k}\sum_{i=1}^{k} L((T_{init}\oplus\delta)\circ T_i^k|I_i,\Theta)
    \label{eq:bid-nerf_opt}
\end{equation}
Eq. \ref{eq:bid-nerf_opt} shows how multiple images can be used during localization. 
It averages the loss function $L$ introduced in Eq. \ref{eq:depth_and_rgb_loss} for the single image use-case, with the only difference that each ray is transformed from the \(i\)-th image's local coordinate frame to the last image's coordinate frame, using their relative transformations \(T_{i}^{k}\). 
\(T_{init}\) is the initial estimate of the absolute pose for the last image $K$.

The use of an image sequence instead of a single image means that we have more viewing angles from which rays can be sampled, with more reference pixel values. We expect this increase in sampling space will greatly improve the performance of the inherent optimization. Also, this increase in information is free from any extra calculations, so it does not reduce execution speed. In \cite{inerf} it was already shown, that rendering more rays improves the performance of the optimization. However, evaluating too many rays is often not feasible on embedded systems. 
We show that we can achieve better results while keeping the computational complexity fixed, namely, equally distributing the same number of sampled rays between the images.
\textit{Our second hypothesis is that by incorporating multiple images while using the same number of rays, the pose estimation will improve in robustness.}

\subsection{Coarse model}
\label{sec:methodology-coarse}
\begin{figure}[h]
    \centering
    \includegraphics[width=\linewidth]{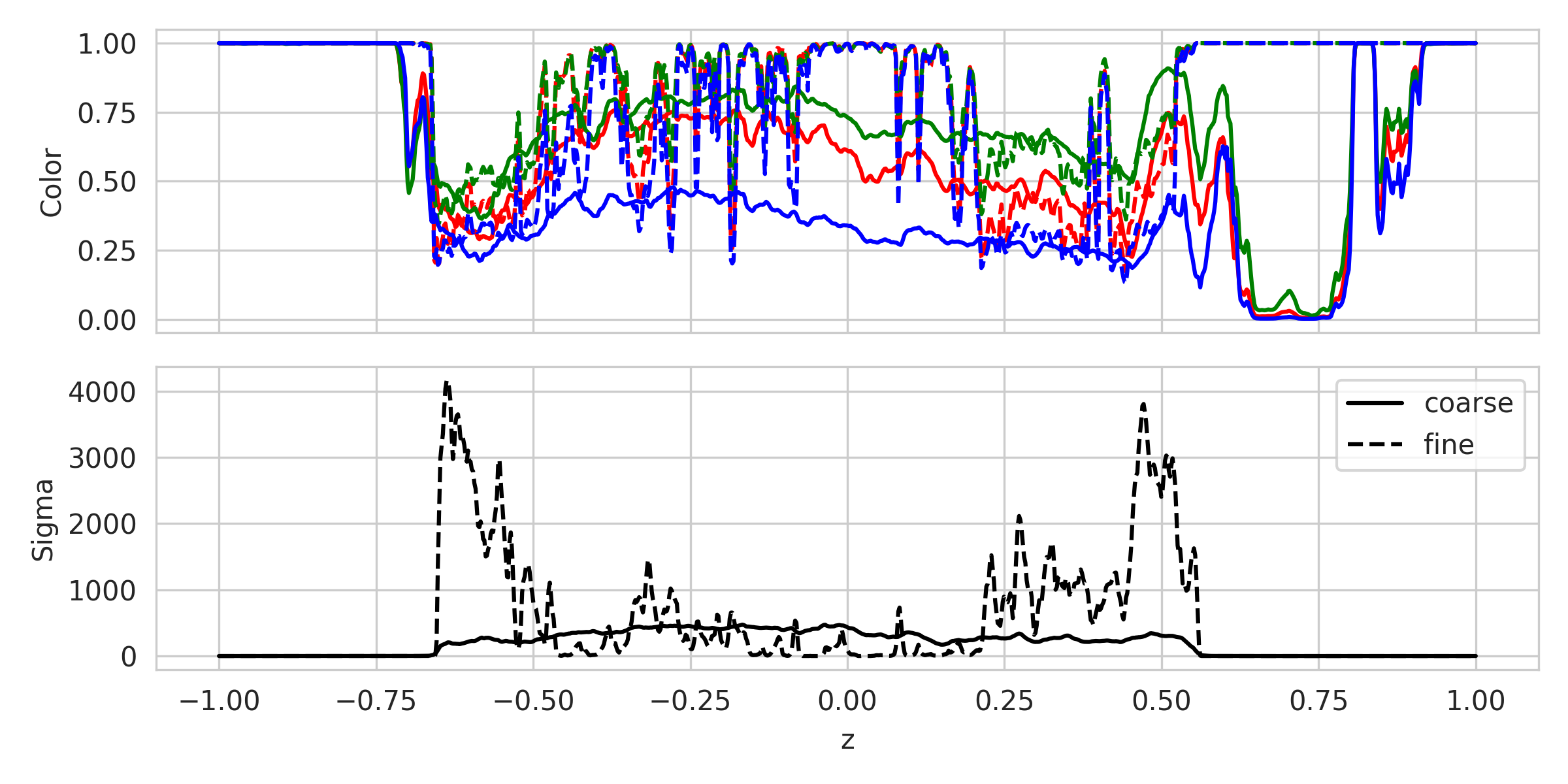}
    \caption{RGB and \(\sigma\) predictions on a ray through the z-axis of the chair object using both the fine and coarse models. Predictions for the fine model are noisier, especially the inside of the objects, where the model is rarely evaluated. \(\sigma\) predictions for the coarse model are flat, and uncertain where the surfaces lie, which is disadvantageous for reconstruction but beneficial for pose optimization.}
    \label{fig:ray_plots_chair}
\end{figure}

\begin{figure}
    \centering
    \begin{subfigure}[b]{.99\linewidth}
        \centering
        \includegraphics[width=\linewidth]{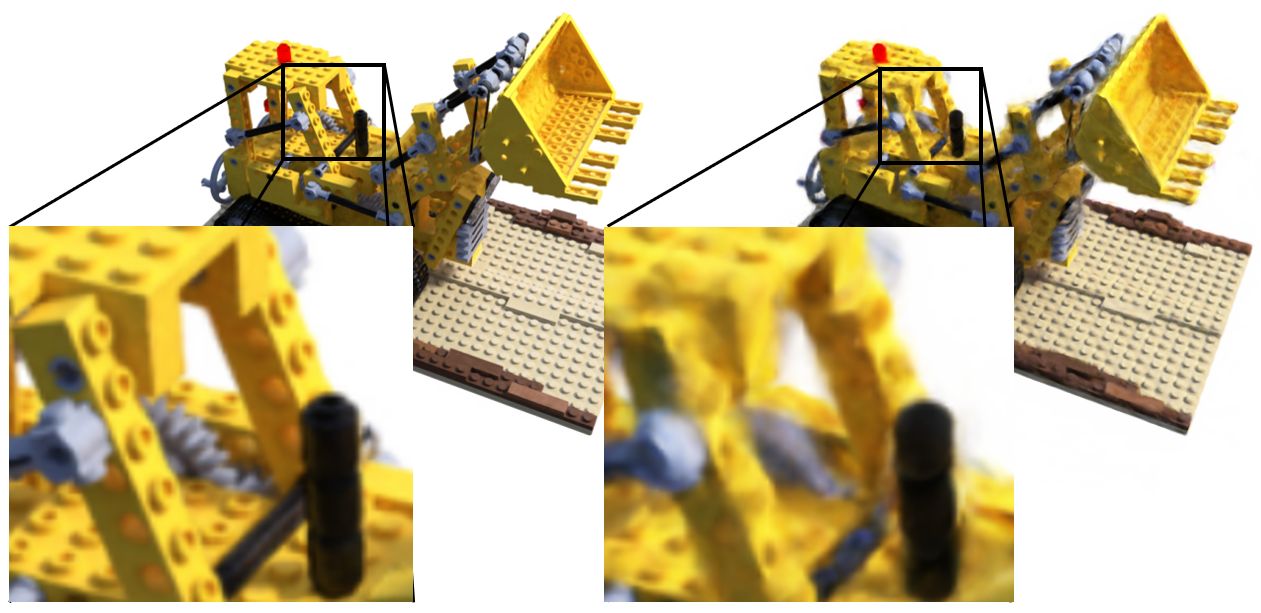}
    \end{subfigure}\\
    \hfill
    \begin{subfigure}[b]{.5\linewidth}
        \centering
        \includegraphics[width=\linewidth]{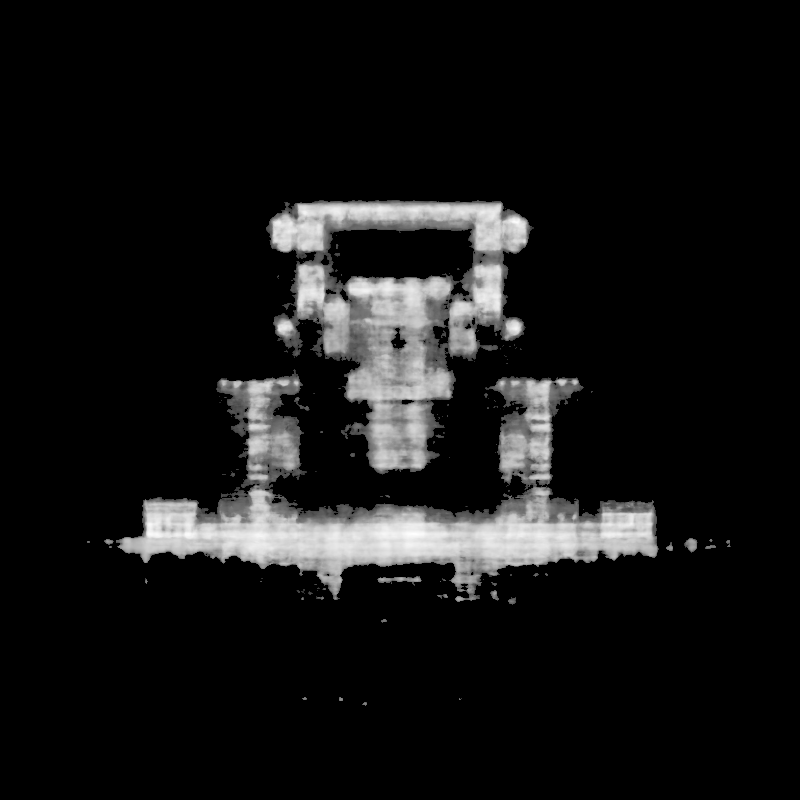}
    \end{subfigure}%
    \hfill
    \begin{subfigure}[b]{.5\linewidth}
        \centering
        \includegraphics[width=\linewidth]{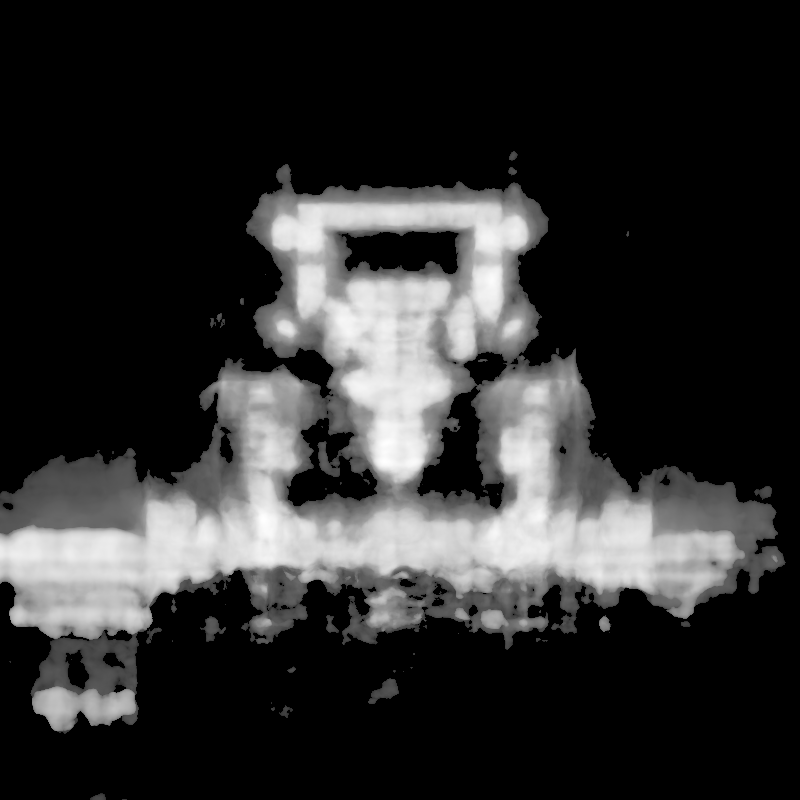}
    \end{subfigure}
    \caption{RGB renderings based on the fine and the coarse model respectively (top) and \(\sigma\) predictions for the Lego object on a vertical slice using the fine and coarse models respectively (bottom).}
    \label{fig:coarse-fine-side-by-side}
\end{figure}

In \cite{nerf} hierarchical sampling for volume rendering was introduced. 
Two separate neural networks: a coarse and a fine model were used for rendering. 
The fine model achieves high accuracy if rays are sampled close to the object's surface.
The purpose of the coarse model is to obtain information about the location of the surfaces in the space, resulting in a more efficient volume sampling method for the fine model. 
This approach improves the reconstruction capabilities of NeRFs, enabling them to represent finer details with higher frequencies. 
Fig. \ref{fig:ray_plots_chair} compares the predictions of the coarse and fine model for both the \(\sigma\) and RGB values along a selected ray. 
Finer details lead to higher image gradients in RGB predictions, as shown on the left side of Fig. \ref{fig:coarse-fine-side-by-side}. 
This effect is undesirable if we want to use the models for pose estimation. 
\textit{Our third hypothesis is that if we purposefully reduce the complexity of the represented space, the localization capabilities of NeRFs will improve.}

We implemented this approach by omitting hierarchical sampling completely and only using the coarse model for pose estimation. This method can be combined with the original hierarchical setup, meaning both coarse and fine models could be used for model training and visualization and only the coarse for localization.


We expect that by using the coarse model for rough pose estimation, the smoother gradients of the RGB and volumetric density values will increase the convergence rate. Fig. \ref{fig:coarse-fine-side-by-side} (bottom) shows the sampled values of the volumetric density for fine and coarse models. It can be observed that the \(\sigma\) predictions of the coarse model are also present in areas away from the surface of the object, while the fine model produces much sharper changes. The gradients of the coarse model's predictions are smoother, making them favorable for first-order optimization algorithms. 
Using only the coarse model helps the convergence and reduces storage requirements if visualization is not necessary, as only one neural network must be stored as a localization map.  
Furthermore, by eliminating the fine model, a substantial decrease in computational complexity can be achieved.
\begin{algorithm}[h]
\caption{BID-NeRF}\label{alg:one}
\SetKwInput{KwInput}{Input}                
\SetKwInput{KwOutput}{Output}              
\DontPrintSemicolon
  \KwInput{$K$ length of image sequence \newline
           $I_i$ i-th RGB-D image and $T_i$ relative pose to the last camera \newline
           $\Theta$ coarse NeRF model \newline
           $\kappa$ intrinsic camera parameters \newline
           $\epsilon$ residual error threshold, $max\_steps$ threshold \newline
           $T_{est}$ initial pose hypothesis for the last image}
  \KwOutput{$T$}
\While{$ s < max\_steps\ and\ L < \epsilon$}{
    \For{$i \in [1..K]$}{
        $uv^{1\dots {m}}_i, c_{i}^{1\dots m}, d_{i}^{1\dots m} = sample\_m\_pixels(I_i)$ \;
        $r^{1\dots m}_{local_i} = cast\_rays(uv^{1\dots m}_i, \kappa)$ \;
        $T^{s}_{abs_i} = (T^{s}_{est}\oplus \delta) \circ T_i$ \;
        $r^{1\dots m}_{model_i} = transform\_rays(r^{1\dots m}_{local_i}, T^{s}_{abs_i})$ \;
        $\hat{c}_{i}^{1\dots m}, \hat{d}_{i}^{1\dots m} = render\_rays(r^{1\dots m}_{model_i}, \Theta)$ \;
        $L_{rgb} = \frac{1}{m}\sum_j\left \| \hat{c}^j_{i}-c^j_{i} \right \|_{2}^{2}$ \;
        $L_{depth} = \frac{1}{m}\sum_{j}\left \| \hat{d}^j_i-d^j_i \right \|_{2}^{2}$ \;
        $L_i = L_{rgb}+\lambda_{d}\cdot L_{depth}$
    }
    $T^{s+1}_{est} = T^{s}_{est} \oplus Adam(\nabla_\delta(\frac{1}{K}\sum L_i))$ \;
    $T = T^{s+1}_{est}$
}
\end{algorithm}
\section{Results}
\begin{figure*}[ht]
    \centering
    \includegraphics[width=\linewidth]{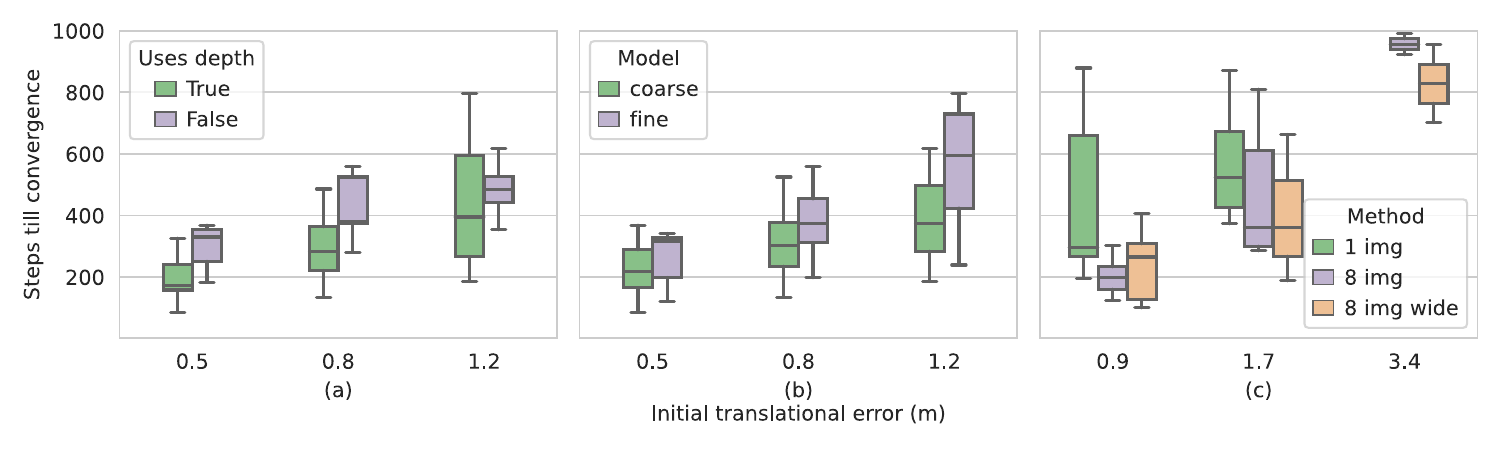}
    \caption{Convergence speed evaluated on all eight objects in the Blender dataset with three different magnitudes of initial translational error. With and without the depth term in the loss function (a), using the coarse or fine models (b), and using one or multiple images (8 images rotated around the object with 15° steps) and ray sampling methods (wide: 2x distance between near and far planes) (c).}
    \label{fig:boxplots}
\end{figure*}

To support the hypotheses we compared the performance of BID-NeRF and iNeRF based on the following performance metrics: convergence speed, accuracy, and robustness based on the Blender dataset \cite{nerf}.
Each test consists of multiple localizations based on RGB-D images with known pose estimations. These initial transformations contain errors in both their rotational and translational components. For the statistical tests, we used perturbation vectors, with fixed lengths and random directions, added to the corresponding elements of the Lie algebra for the rotational and translational components. In \cite{inerf}, translational perturbation vectors with lengths within [0, 0.2] meters were used. To show that the range of convergence can be expanded, we used vectors with lengths between [0.5, 3.4] meters. We show the length of the translational perturbation vector in meters where the reference objects are scaled into a 2x2x2m cube. The lengths of the rotational perturbation vectors were kept in a range of [0.2, 1.4], which resulted in rotational errors under absolute 45 degrees, measured between the ground truth and the estimated rotational translations. 

Each optimization is run for 1000 steps. For each experiment, we use the same amount of rays (2048) distributed between different input images. We used Huber loss \cite{10.1214/aoms/1177703732} as loss function to decrease the impact of outliers. Pixels are resampled for each iteration as it adds further stochasticity to the optimization. First-order optimization algorithms often benefit from stochastic gradient estimation as it helps them to avoid getting stuck in local minima. 

The pose estimation is considered successful if the original translational error is reduced to at least 10\%.
The convergence speed is measured in the optimization steps needed to achieve this threshold. In real applications the optimal transformation is unknown. In these scenarios, a meaningful heuristic could be used based on the final loss function to detect convergence, as demonstrated in Alg. \ref{alg:one}.

\subsection{Depth information and Coarse model}

As shown in Fig. \ref{fig:boxplots} (a-b), convergence speed is increased by the incorporation of depth information and utilization of the coarse model instead of the fine. 
These results only contain the tests where convergence was achieved. 
Table \ref{tab:coarse-fine-depth-convergence} shows the percentage of convergence combined for all initial translational error magnitudes. Compared to the original iNeRF approach by applying the the depth regularization term and by using only the coarse model a 70\% increase in the number of converged tests is achieved.
\begin{table}[h]
    \centering
    \begin{tabular}{l|cc}
         & Fine & Coarse\\
        \hline
        RGB + Depth & 87.5\% & \textbf{95.83\%} \\
        RGB & 25\% & 75\%
    \end{tabular}
    \caption{Percentage of converged tests for the Blender dataset with three different initial translational errors combined [0.5m, 0.8m, 1.2m].}
    \label{tab:coarse-fine-depth-convergence}
\end{table}

Depth information has already been proven useful for localization tasks in many applications. The reason why using the coarse model has similar effects might seem unintuitive. During the training of NeRFs, the coarse model is evaluated at random points in the space with a uniform distribution. In contrast, the fine model is more likely to be evaluated closer to the surfaces. This makes the rendered images more accurate compared to the images rendered with the coarse model as seen in Fig. \ref{fig:coarse-fine-side-by-side} (top), which seems to be more blurred. The same can be said about the raw \(\sigma\) predictions of the model, shown in Fig. \ref{fig:coarse-fine-side-by-side} (bottom). In line with our previous hypothesis in section \ref{sec:methodology-coarse}, these blurred predictions seem to help the optimization, achieving the same effect as applying a low pass filter to the image.
\subsection{Multiple images}
Image pose estimation based on image sequences can be easily simulated with the Blender dataset. Every model has 200 reference images with known poses around the represented object(s). We took a subset of these images, calculated their relative transformations, and perturbed their absolute pose together around the last image, as seen in Fig. \ref{fig:multi}. We experimented with the distance between these images and found that at small distances the algorithm benefitted little from the incorporation of previous frames. This comes from the fact that there is little to no difference between images next to each other. 
In order to reduce the visual similarity we decided to select frames around the reference object based on distance and orientation changes.
The further these images lie from each other, the better coverage of the space can be achieved.
On the other hand, by increasing the distance between the images, we attenuate our assumption that the errors of their relative poses are insignificant compared to the error in their absolute pose estimate.
The optimal choice of distance is application-dependent.


\begin{table}[h]
    \centering
    \begin{tabular}{l|ccc}
         &\multicolumn{3}{|c}{Initial translational error}\\
         & 0.9m & 1.7m & 3.4m\\
        \hline
        1 image & 87.5\% & 50\% & 0\%\\
        8 images & 87.5\% & 87.5\% & 25\%\\
        8 images + wide & 87.5\% & 75\% & 25\%\\
    \end{tabular}
    \caption{Percentage of converged tests for the Blender dataset with three different initial translational errors using different pixel sampling methods.}
    \label{tab:multi-convergence}
\end{table}

The idea of sampling rays from more than one virtual camera widens the range of the space where similarities are searched for. 
Fig. \ref{fig:boxplots} (c) and Table \ref{tab:multi-convergence} show the results of quantitative experiments of this modification. 
The incorporation of multiple input images seems beneficial when using images that are farther from each other (15° in degrees where every camera is facing toward the object). At higher initial errors we see a difference in convergence percentage which shows that by using more images we can widen the basin of convergence, leading to a more robust optimization algorithm. Using a broader sampling space along the rays improves the speed of convergence in these tests.
\subsection{BID-NeRF vs iNeRF}


\begin{figure}[h]
  \centering
  \includegraphics[width=0.8\linewidth]{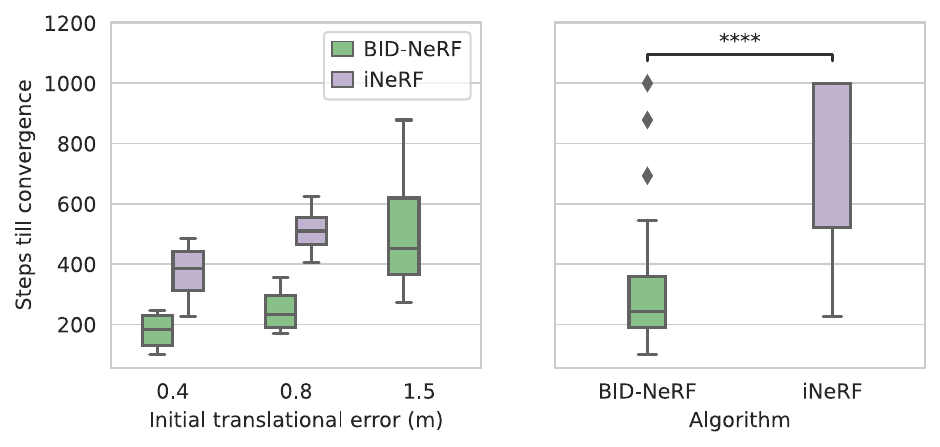}
  \caption{Statistical analysis of convergence speed. All eight objects are evaluated in the Blender dataset with three different initial translational error sizes (left). Only the converged experiments are included.\\ Combined results with statistical Mann-Whitney U test \cite{10.1214/aoms/1177730491} to check significance (right). Four * means that the p-value is below $10^{-4}$.}
  \label{fig:bidnerf_inerf_boxplots}
\end{figure}

\begin{figure}[H]
    \centering
    \includegraphics[width=\linewidth]{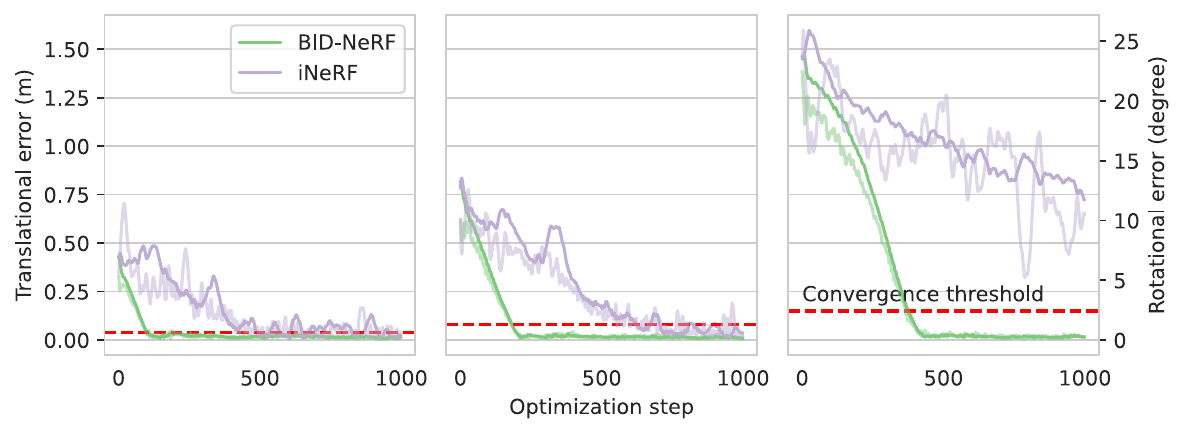}
    \caption{Rotational (faded) and translational errors throughout optimizations based on the chair object from three different initial translational errors ([0.4, 0.8, 1.5] respectively).}
    \label{fig:bidnerf-inerf_steps}
\end{figure}

Finally, we compare the iNeRF algorithm to BID-NeRF, where all our proposals mentioned in Sec. \ref{methodology} are incorporated.
The statistical analysis of convergence speed for all eight models from the Blender dataset with three different magnitudes of initial translational error is shown in Fig. \ref{fig:bidnerf_inerf_boxplots}.
Fig. \ref{fig:bidnerf-inerf_steps} shows how the translational and rotational error decreases during optimization for the chair object.
BID-NeRF vastly outperforms the iNeRF algorithm, especially for larger initial errors. 

\begin{figure*}[!h]
    \centering
    \includegraphics[width=\linewidth]{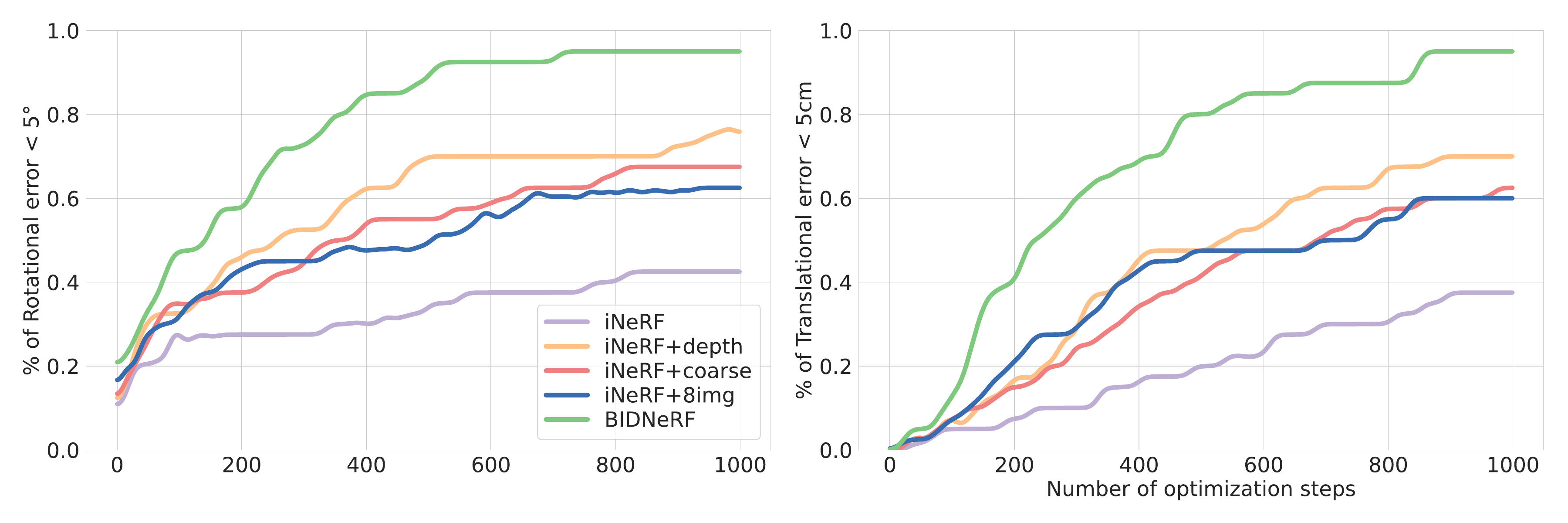}
    \caption{Ablation study showing the ratio of converged tests after each optimization step. 5 pose estimations were evaluated from different poses for each object in the Blender dataset. The initial error formulation is identical to the test setup presented in \cite{inerf}.}
    \label{fig:cumulative}
\end{figure*}
Fig. \ref{fig:cumulative} shows an ablation study, where most of our modifications yielded significant improvements in terms of optimization steps needed for convergence while also increasing the domain of convergence.
The configuration of the experiments with regard to: the initial pose error, learning rate, and pixel count was identical to the ones used in \cite{inerf}.
BID-NeRF achieved 95\% convergence in both error types after 1000 steps, while this percentage is only at 38\% for translational and 42\% for rotational errors using the iNeRF algorithm.



\subsection{Real-time performance}

Our proposed solution is NeRF model agnostic, in the sense that it can be applied to any differentiable NeRF model.
For example, \cite{kurz2022adanerf} is one of the fastest currently available NeRF implementations, which renders an image with a resolution of 1008×756 pixels of an LLFF object in 38-130ms depending on the desired quality using an Nvidia RTX 3090 GPU.
They achieved this result by using a depth predictor network instead of the hierarchical sampling approach, significantly reducing the number of forward passes in inference time.
Our method performs 2048 forward and backward evaluations to calculate the color and depth predictions and gradients for the input poses for every iteration.
This process only takes 8ms using the PyTorch implementation of AdaNeRF. (The PyTorch implementation achieves a significantly lower performance compared to their real-time CUDA-based renderer.)
For the Blender dataset, our method converges in 600-800 iterations
for very high initial errors, resulting in a runtime of 5-6s with AdaNeRF.
In \cite{lin2023parallel} the same process with lower initial errors takes 15-20s on an RTX 3090, and in \cite{maggio2022locnerf} 200 update steps are required for convergence, where one update step is 0.6 Hz for 300 particles and 1.8 Hz for 100 particles on an RTX 5000 GPU.
Our strategy of not using a particle filter framework and using a better optimization objective pays off making BID-NeRF superior compared to \cite{lin2023parallel, maggio2022locnerf} from a computational complexity perspective.

It is important to note, that pose estimation for drift removal does not have to be evaluated for every image.
If our method is used for continuous tracking, the errors in the initial pose estimates are usually much smaller, resulting in faster convergence.

\section{Conclusion}
Our proposed method BID-NeRF outperformed iNeRF in image pose estimation by achieving 2-4 times improvement in the speed of convergence, depending on the size of initial perturbation, and improved robustness meaning that successful convergence rate was 2.5 times better on the Blender dataset.
With the elimination of hierarchical sampling the inference time of every optimization step was reduced by a factor of two, together with the storage requirements of the algorithm. 
The proposed extensions such as sequences of images with estimated relative poses and sparse depth information are usually available in robotic systems, thus there is no need for extra computational effort to produce the required data.

{\small
\bibliographystyle{ieee_fullname}
\bibliography{egbib}
}

\end{document}